\documentclass[10pt]{article}

\usepackage[margin=0.82in]{geometry}
\usepackage{times}
\usepackage{microtype}
\usepackage{graphicx}
\usepackage{amsmath,amssymb}
\usepackage{booktabs}
\usepackage{multirow}
\usepackage{array}
\usepackage{tabularx}
\usepackage{longtable}
\usepackage[table]{xcolor}
\usepackage[numbers,sort&compress]{natbib}
\usepackage[colorlinks=true,linkcolor=blue!55!black,citecolor=blue!55!black,urlcolor=blue!55!black]{hyperref}
\usepackage{caption}
\usepackage{subcaption}
\usepackage{enumitem}
\usepackage{titlesec}
\usepackage{placeins}
\usepackage{url}
\usepackage{makecell}
\usepackage{threeparttable}
\usepackage{etoolbox}
\usepackage{fontawesome5}

\definecolor{linkblue}{HTML}{2C6CB0}
\definecolor{linkgold}{HTML}{C68B16}

\newcommand{\codeurl}{https://github.com/rdharini2001/ICIP_2025_Data_Release}
\newcommand{\dataurl}{https://huggingface.co/datasets/Dharini24/AVID}
\newcommand{\siteurl}{https://rdharini2001.github.io/ICIP_2025_Data_Release/}

\graphicspath{{figures/}}
\setlist[itemize]{leftmargin=1.35em,itemsep=1.5pt,topsep=3pt}
\setlist[enumerate]{leftmargin=1.55em,itemsep=1.5pt,topsep=3pt}
\titleformat{\section}{\large\bfseries}{\thesection}{0.7em}{}
\titleformat{\subsection}{\normalsize\bfseries}{\thesubsection}{0.7em}{}
\titleformat{\subsubsection}{\normalsize\bfseries}{\thesubsubsection}{0.6em}{}
\pretocmd{\section}{\FloatBarrier}{}{}

\newcommand{\dataset}{\textsc{AVID}}
\newcommand{\corrupt}{\textsc{AVID-C}}
\newcommand{\rgb}{\textsc{RGB}}
\newcommand{\ir}{\textsc{IR}}
\newcommand{\Bhat}{\widehat B}
\newcommand{\bfg}{\Delta_{\mathrm{fusion}}}

\newcommand{\burden}{\widehat A}

\hypersetup{
  pdftitle={AVID: A Weakly Aligned RGB-Infrared Dataset and Perturbation Benchmark for Tiny Aerial Targets},
  pdfauthor={Dharini Raghavan and Vishnu Rao and Sethu Selvi S},
  pdfsubject={Tiny aerial-target perception with weakly aligned RGB-infrared images and simulated perturbations},
  pdfkeywords={RGB-infrared, thermal, multimodal, drone, bird, tiny aerial targets, perturbation benchmark, weak alignment, dataset}
}

\title{
\vspace{-1.2em}
\textbf{AVID: A Weakly Aligned RGB-Infrared Dataset and
Perturbation Benchmark for Tiny Aerial Targets}
}

\author{
Dharini Raghavan$^{1,*}$ \quad
Vishnu Rao$^{2,*}$ \quad
Sethu Selvi S$^{3}$\\[5pt]
\small
$^1$Georgia Institute of Technology \quad
$^2$ARTPARK, Indian Institute of Science\\[2pt]
\small
$^3$Ramaiah Institute of Technology\\[3pt]
\small
\texttt{draghavan7@gatech.edu}
}

\date{}

\begin{document}

\maketitle

\begingroup
\renewcommand{\thefootnote}{\fnsymbol{footnote}}
\footnotetext[1]{Dharini Raghavan and Vishnu Rao contributed equally to this work.}
\endgroup

\vspace{-0.7em}
\begin{center}
\small
{\color{linkblue}\faGithub}\;
\href{\codeurl}{Code}
\qquad
{\color{linkgold}\faSmileBeam}\;
\href{\dataurl}{Hugging Face}
\qquad
\href{\siteurl}{Research website}
\end{center}
\vspace{-0.3em}

\begin{abstract}
Long-range detection of small drones is complicated by their visual similarity to birds and by the limited detail available from ground-based cameras. Combining visible and infrared observations offers complementary information, but their usefulness depends on target size, cross-modal alignment, and image quality. We present \dataset{}, a benchmark for studying these factors using real RGB-infrared bird-drone surveillance footage originating from the Drone Detection Dataset and curated for the IEEE ICIP 2025 Video and Image Processing Cup. \dataset{} comprises 32,040 corresponding frame pairs from 115 videos, with 35,952 existing bird and drone annotations at $320\times256$ resolution. Targets occupy an unusually narrow size range: 68.1\% have a geometric-mean bounding-box side of 8--16 pixels, and 99.0\% meet the COCO small-object criterion. To characterize the geometric difficulty of combining the modalities, we introduce an object-relative alignment ratio and an alignment burden curve. In a sample of 300 frame pairs, 32.3\% have estimated global translations greater than 5 pixels; among the 264 pairs with matching annotations, 16.3\% exceed one median target side. We further introduce \corrupt{}, which simulates seven families of image degradation at five severity levels independently in RGB, infrared, or both. A fixed-feature analysis examines how scale, degradation, and added cross-modal shifts affect fitted bird-drone class overlap. At 12--16 pixels, the Bhattacharyya error bounds are 0.148 for RGB, 0.198 for infrared, and 0.046 for their combined features. These values describe separation in the specified feature and density models, not detector accuracy. Together, \dataset{} and \corrupt{} support scale-aware, alignment-aware, and sensor-specific studies of tiny aerial targets.

\end{abstract}

\section{Introduction}

A distant drone may occupy only a few pixels in a ground-based surveillance image. At this scale, the silhouette and texture that distinguish a drone from a bird are often poorly resolved, while background clutter, blur, and sensor noise can dominate the target appearance. Recognition is therefore difficult not only because the objects are small, but because their class-specific visual information is easily lost. Prior work has addressed bird-drone classification \citep{selvi2022deep}, drone localization and tracking \citep{10193217,10328384}, and multimodal drone localization \citep{11700652}. A complementary question concerns the observations themselves: how much class information remains when targets are tiny, the camera views are displaced, or one sensor is degraded?

Visible and infrared imagery offer different views of the same aerial target. RGB captures appearance and scene context, whereas infrared records thermal contrast that may persist under poor visible-light conditions. These benefits are not unconditional. The cameras have different optics and response characteristics, and corresponding frames need not be registered. A shift of five pixels is minor for a large object but substantial for a target whose bounding box is ten pixels wide. Likewise, a perturbation affecting one modality may have a different effect on the other. Consequently, evaluations that report only an aggregate detection score can obscure the conditions under which multisensor observations become useful or unreliable.

\dataset{} provides a setting for studying these questions with birds and drones observed by ground-based RGB and infrared cameras. The benchmark comprises 32,040 corresponding frame pairs from 115 surveillance videos and 35,952 annotated instances. Its footage originates from the Drone Detection Dataset of Svanstr\"om and colleagues \citep{svanstrom2021multisensor,svanstrom2020repository}, with bird-drone observations curated for the IEEE ICIP 2025 Video and Image Processing Cup \citep{ieeevipcup2025}. Importantly, correspondence between the RGB and infrared views does not imply pixel-level registration. Rather than treating this discrepancy as incidental, \dataset{} makes it part of the evaluation setting.

Two properties distinguish the evaluation setting. First, 24,489 annotations (68.1\%) have a geometric-mean bounding-box side of only 8--16 pixels in the native $320\times256$ images; 99.0\% belong to the COCO small-object category. We therefore examine finer size intervals rather than treating virtually all targets as a single size group. Second, sensor displacement must be interpreted at the scale of these objects. We define an object-relative alignment ratio and an alignment burden curve, and estimate global RGB-infrared shifts in a 300-pair sample. Among 264 annotation-matched pairs, 16.3\% have an estimated shift exceeding one median target side.

To study image degradation separately from target geometry, we introduce \corrupt{}, a controlled benchmark comprising seven perturbation families at five severity levels. Noise, blur, illumination changes, and camera instability can be simulated in either modality or in both, making the affected sensor an explicit evaluation variable. We additionally analyze bird-drone separation in a fixed low-dimensional feature space under changes in target size, image quality, and cross-modal shift. This analysis is conditional on annotated target locations and does not substitute for evaluating trained detectors.

The contributions are as follows:
\begin{enumerate}
    \item \textbf{A paired benchmark for tiny bird-drone surveillance targets.} \dataset{} provides 32,040 corresponding RGB-infrared frames from 115 videos, existing bounding-box annotations, and sequence identifiers suitable for video-disjoint evaluation.
    \item \textbf{Target-relative alignment measures.} An object-relative displacement ratio, burden curve, and summary score describe estimated camera shifts in units of target size, rather than in pixels alone.
    \item \textbf{Sensor-specific robustness testing.} \corrupt{} defines seven simulated image-degradation families at five severity levels for RGB-only, infrared-only, and joint perturbation experiments.
    \item \textbf{An analysis of scale, degradation, and alignment.} Fixed-feature class-overlap estimates characterize these factors and include controls for the effect of representation dimension.
\end{enumerate}

\section{Related datasets and benchmark design}
\label{sec:related}

\subsection{Visible-infrared and UAV datasets}

Visible-infrared benchmarks have established the value of complementary sensors across several perception tasks. KAIST introduced synchronized RGB-thermal imagery for pedestrian detection \citep{hwang2015kaist}; LLVIP studies paired visible-infrared observations in low-light scenes \citep{jia2021llvip}. M3FD extends multimodal fusion and detection to varied environments \citep{liu2022tardal}, while FLIR ADAS focuses on thermal and visible perception in driving scenes \citep{flir2018adas}. These datasets primarily emphasize ground-level objects with imaging conditions distinct from distant bird-drone discrimination.

UAV benchmarks differ both in the target and in the camera viewpoint. VisDrone concerns objects viewed from airborne cameras \citep{zhu2022visdrone}, whereas \dataset{} concerns an aerial target viewed from the ground. DroneRGBT addresses aerial crowd counting \citep{peng2020dronergbt}, and VTUAV introduces paired visible-thermal sequences for tracking \citep{zhang2022vtuav}. Anti-UAV directly addresses drone surveillance and provides visible and infrared observations \citep{jiang2023antiuav,antiuavrepo}. Related methods have studied bird-drone recognition \citep{selvi2022deep}, drone localization and tracking \citep{10193217,10328384}, and sensor fusion for localization \citep{11700652}.

Recent UAV datasets further broaden sensing conditions and recognition tasks. MMFW-UAV emphasizes multi-view, multi-sensor fixed-wing observations \citep{liu2025mmfwuav}; MM-UAV combines RGB, infrared, and event data \citep{xu2025mmuav}; UAVNet-MS addresses fine-grained recognition with RGB and multispectral imagery \citep{luo2026uavnetms}; and UAV-CB considers complex backgrounds and camouflage \citep{huang2026uavcb}. These resources complement rather than replace a characterization of target scale, residual cross-modal translation, and sensor-specific degradation in ground-based bird-drone footage.

The surveillance imagery underlying \dataset{} was collected by Svanstr\"om et al. \citep{svanstrom2021multisensor} and the bird-drone observations were curated for the VIP Cup. Within this setting, \dataset{} emphasizes three factors that are difficult to disentangle using conventional benchmark scores: very small target size, residual RGB-infrared displacement, and sensor-specific image degradation. Table~\ref{tab:related} situates this focus among related datasets. As in the rest of the paper, paired observations should not be confused with pixel-registered images.

\begin{table}[t]
\centering
\caption{\textbf{Comparison of \dataset{} with related visible-infrared and UAV datasets.}
The pairing column describes the correspondence provided between modalities. Paired or synchronized observations do not necessarily imply pixel-level alignment.}
\label{tab:related}
\small
\begin{threeparttable}
\begin{tabularx}{\linewidth}{@{}l l l l X@{}}
\toprule
Dataset & Main target/view & Modalities & Pairing & Main characteristic \\
\midrule

KAIST \citep{hwang2015kaist}
& pedestrian, ground
& RGB + thermal
& synchronized
& multispectral pedestrian detection \\

LLVIP \citep{jia2021llvip}
& pedestrian, ground
& visible + IR
& paired
& low-light pedestrian imagery \\

M3FD \citep{liu2022tardal}
& multi-class, ground
& visible + IR
& paired
& image fusion and object detection \\

VTUAV \citep{zhang2022vtuav}
& generic target, aerial
& visible + thermal
& paired video
& visible-thermal object tracking \\

Anti-UAV300 \citep{antiuavrepo}
& UAV, surveillance
& RGB + IR
& both streams released
& long-range UAV surveillance \\

MMFW-UAV \citep{liu2025mmfwuav}
& fixed-wing UAV, air-to-air
& visible + thermal
& one-to-one
& multi-sensor and multi-view observations \\

MM-UAV \citep{xu2025mmuav}
& UAV tracking
& RGB + IR + event
& synchronized
& tracking with three sensor modalities \\

UAVNet-MS \citep{luo2026uavnetms}
& small UAV
& RGB + MSI
& synchronized
& multispectral UAV recognition \\

UAV-CB \citep{huang2026uavcb}
& UAV, low-altitude
& RGB + thermal
& paired
& UAV detection in complex backgrounds \\

\rowcolor{blue!4}
\dataset{} (ours)
& bird/drone, surveillance
& RGB + thermal
& weakly aligned pairs
& tiny targets, measured alignment, and controlled perturbations \\

\bottomrule
\end{tabularx}
\end{threeparttable}
\end{table}

\subsection{Corruption and data-centric benchmarks}

Robustness benchmarks make image degradation an explicit experimental variable. ImageNet-C organizes common corruptions by type and severity \citep{hendrycks2019benchmarking}, and UAV-C extends corruption-based evaluation to aerial tracking \citep{liu2024uavc}. MapBench similarly studies environmental and sensor disturbances in multimodal mapping \citep{hao2024mapbench}. Their central lesson is that averaging across perturbations can conceal the failure modes of individual conditions.

\corrupt{} applies this principle to low-resolution paired surveillance. Its seven perturbation families are parameterized for the native $320\times256$ images and can affect RGB, infrared, or both. In contrast to a single pooled robustness score, this design exposes which modality is degraded and by how much. The benchmark does not assume that nominal perturbation severity is equivalent to downstream task difficulty.

Data-centric evaluation also motivates separating properties of a dataset from properties of a trained model. DataPerf and DataComp examine how data quality, selection, and composition affect learning \citep{mazumder2023dataperf,gadre2023datacomp}. WhyShift emphasizes identifying the sources of distribution change \citep{liu2023whyshift}, while Mayo et al. investigate image difficulty beyond a single model score \citep{mayo2023difficulty}. Following this line of work, we characterize object size, spatial correspondence, and simulated degradation before interpreting class separation in a fixed representation.

\section{The AVID dataset}
\label{sec:dataset}

\subsection{Observation setting and provenance}

\dataset{} consists of RGB and infrared observations of birds and drones captured by ground-based surveillance cameras. The two modalities provide complementary views of distant aerial targets, which frequently occupy only a small region of the image. The imagery originates from the Drone Detection Dataset of Svanstr\"om, Alonso-Fernandez, and Englund \citep{svanstrom2021multisensor,svanstrom2020repository}; the bird-drone subset was curated for the IEEE ICIP 2025 Video and Image Processing Cup \citep{ieeevipcup2025}. The benchmark focuses on bird-drone observations rather than the challenge's separate payload-identification task.

Each RGB-infrared pair represents corresponding observations of a scene. Because the cameras have separate optics, the target may appear at different image coordinates in the two views. Pairing identifies corresponding observations, not pixel-level registration. Sequence-level identities support evaluation protocols that avoid placing adjacent frames in both training and test sets.

\subsection{Composition and annotations}

\dataset{} contains 32,040 paired frames from 115 videos at a native resolution of $320\times256$ pixels. The Train partition comprises 28,790 pairs from 102 videos, and the Validation partition contains 3,250 pairs from 13 different videos (Table~\ref{tab:corpus}). In total, 76 sequences depict drones and 39 depict birds. Clip lengths range from 50 to 329 paired frames, with a median of 306 and a mean of 278.6.

\begin{figure}[t]
    \centering
    \includegraphics[width=0.67\linewidth]{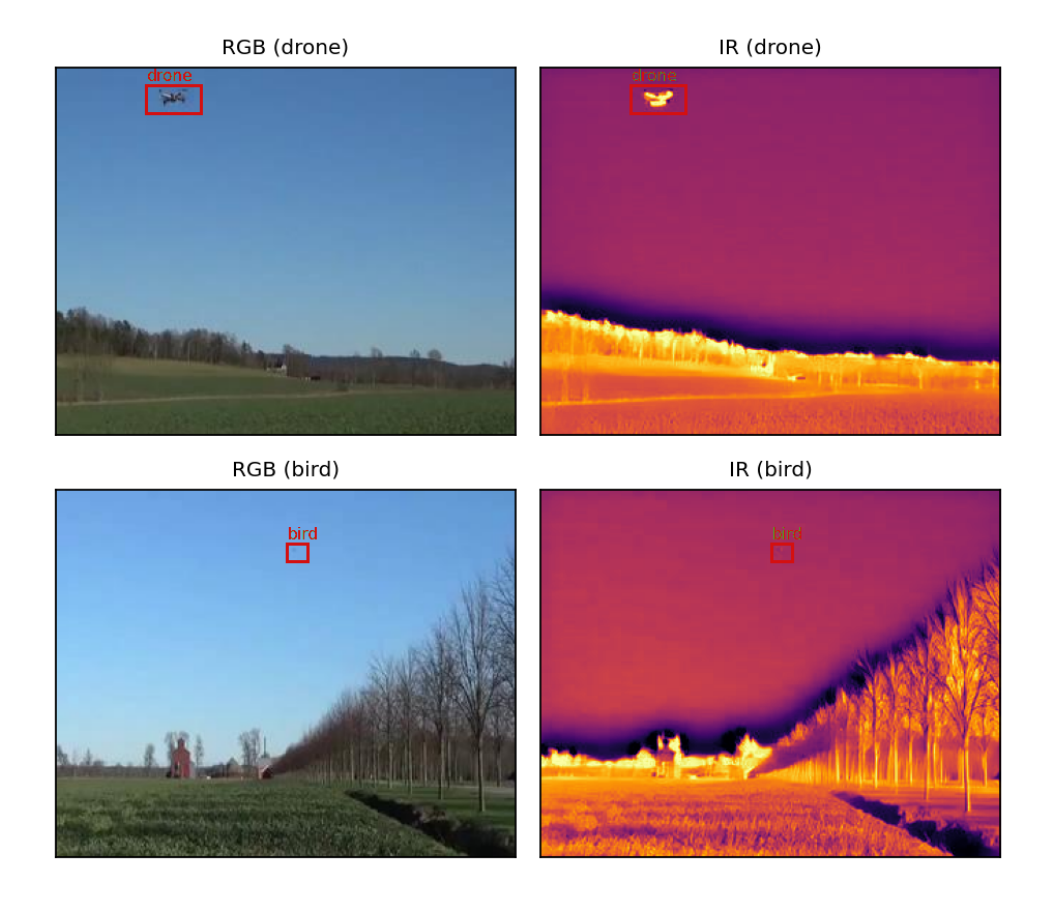}
    \caption{\textbf{Representative RGB-infrared frame pairs from \dataset{}.}
    The top row shows a drone, and the bottom row shows a bird. Red boxes mark the annotated targets. RGB and infrared provide corresponding observations but differ in target appearance and can retain spatial displacement.}
    \label{fig:realpairs}
\end{figure}

The dataset includes 35,952 annotated targets: 15,471 birds and 20,481 drones. The bounding boxes and class labels originate from the VIP Cup curation. They enable object-level analyses of target scale and appearance, while video-level grouping supports sequence-disjoint evaluation. Figure~\ref{fig:realpairs} illustrates representative examples. The different visual and thermal appearances of the same small target, together with residual positional differences between the views, motivate the geometry and perturbation analyses that follow.

\begin{table}[t]
\centering
\caption{\textbf{Composition of \dataset{}.}
Train and Validation contain disjoint source videos. Instance counts refer to the existing bird and drone bounding-box annotations.}
\label{tab:corpus}
\small
\begin{tabular}{lrrrrr}
\toprule
Split & RGB-IR pairs & Videos & Bird instances & Drone instances & Total instances \\
\midrule
Train      & 28,790 & 102 & 14,113 & 18,335 & 32,448 \\
Validation & 3,250  & 13  & 1,358  & 2,146  & 3,504 \\
\midrule
Total      & 32,040 & 115 & 15,471 & 20,481 & 35,952 \\
\bottomrule
\end{tabular}
\end{table}

\section{The geometry of the dataset}
\label{sec:geometry}

\subsection{A concentrated tiny-object regime}
\label{sec:scale}

Target scale is a defining property of \dataset{}. At long range, birds and drones are represented by few pixels, limiting the available shape and texture cues. We quantify this regime using the native bounding-box annotations rather than the resized inputs of a particular model.

Under the COCO area convention \citep{lin2014coco}, 35,581 of 35,952 instances (99.0\%) are ``small,'' 371 are ``medium,'' and none are ``large.'' This concentration makes the usual small/medium/large partition inadequate for distinguishing levels of difficulty within \dataset{}.

We instead characterize each box by its geometric-mean side length,

\begin{equation}
s=\sqrt{w_{\mathrm{px}}h_{\mathrm{px}}},
\end{equation}

where $w_{\mathrm{px}}$ and $h_{\mathrm{px}}$ denote box width and height in native pixels. This quantity equals the side length of a square with the same area as the annotated box.

Figure~\ref{fig:datasetstats}a shows that 17,706 instances fall in the 12--16 pixel interval. Together, the 8--12 and 12--16 pixel intervals contain 24,489 instances (68.1\%). Only 103 instances have sides below 8 pixels, and none exceed 48 pixels. Finer intervals are therefore necessary to assess changes within the overwhelmingly small-object distribution.

The scale distribution also changes the interpretation of sensor displacement. A translation of three to five pixels can span a substantial fraction of the target, even when it appears small relative to the full $320\times256$ image. This motivates expressing alignment in target-side units.

\subsection{Weak RGB-infrared alignment}
\label{sec:alignment}

Corresponding RGB and infrared frames share a scene but are not guaranteed to be pixel-registered. Separate optics and viewpoints can place the same bird or drone at different image coordinates. We estimate the dominant displacement and then evaluate its magnitude relative to object size.

The global translation is estimated by maximizing normalized mutual information (NMI) between each grayscale-converted RGB frame and its infrared counterpart. Integer horizontal and vertical offsets are searched over $[-12,12]$ pixels. Additional estimator details are provided in Appendix~\ref{app:alignment}.

An absolute shift is not equally consequential at different scales: five pixels correspond to half the side length of a 10-pixel target but only one tenth of a 50-pixel target. We therefore normalize each estimated displacement by the target scale observed in that frame.

For pair $i$, let $d_i$ denote the magnitude of the estimated global translation and $s_i$ the median geometric-mean side of its annotated targets. The object-relative alignment ratio is

\begin{equation}
\rho_i=\frac{d_i}{s_i}.
\label{eq:rho}
\end{equation}

Thus, $\rho_i=0.5$ indicates a displacement of half the median target side, and $\rho_i>1$ indicates a shift larger than that side. The empirical alignment burden curve summarizes their prevalence across annotation-matched pairs:

\begin{equation}
\burden(\tau)=\frac{1}{n}\sum_{i=1}^{n}\mathbf{1}[\rho_i>\tau].
\label{eq:burden}
\end{equation}

Here $n$ is the number of pairs with matching annotations, and $\mathbf{1}[\cdot]$ is the indicator function. The curve gives the proportion of pairs exceeding a target-relative threshold $\tau$; in particular, $\burden(0.5)$ is the proportion exceeding half a median target side.

A scalar summary is obtained by integrating the burden curve up to a threshold $T$:

\begin{equation}
\widehat{\mathcal B}_T=\frac{1}{T}\int_0^T\burden(\tau)\,d\tau
=\frac{1}{nT}\sum_{i=1}^n\min(\rho_i,T).
\label{eq:burdenscore}
\end{equation}

For $T=1$, the score is the mean object-relative shift after values above one target side are clipped to one. The scalar facilitates comparison, while the curve retains the distribution of exceedance probabilities.

\begin{figure}[t]
    \centering
    \includegraphics[width=0.99\linewidth]{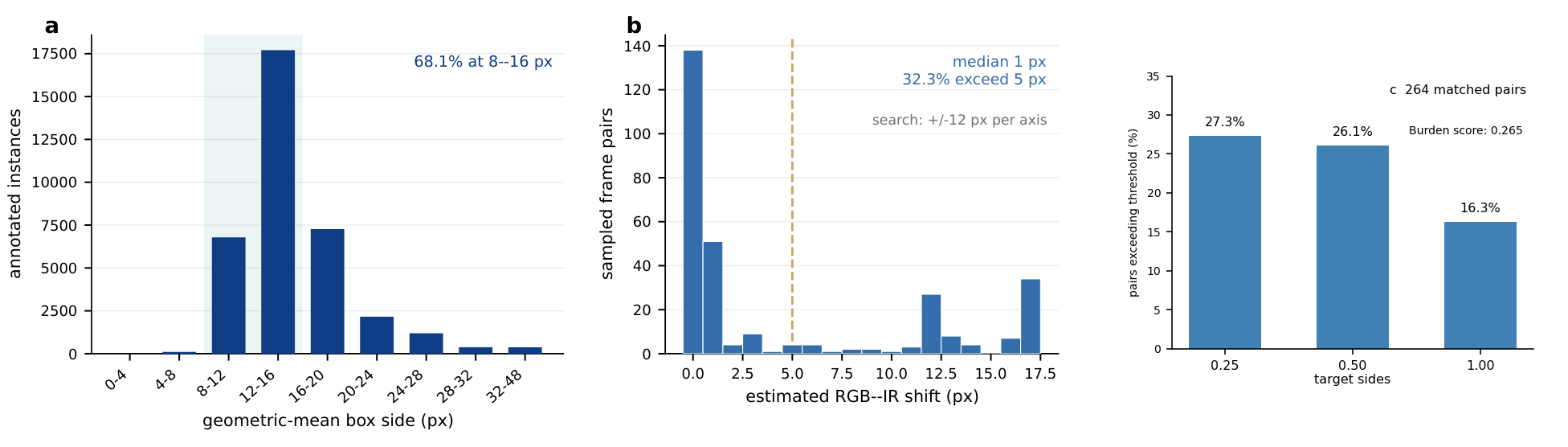}
    \caption{\textbf{Target size and RGB-infrared alignment in \dataset{}.}
    \textbf{a}, Distribution of geometric-mean target side lengths across the Train and Validation data. The highlighted 8--16 pixel range contains 68.1\% of the annotated instances.
    \textbf{b}, Estimated RGB-infrared translation magnitude for 300 sampled frame pairs. The search is limited to $\pm12$ pixels along each axis, corresponding to a maximum measurable translation magnitude of 16.97 pixels.
    \textbf{c}, Alignment burden for the 264 pairs with matching target annotations. The fractions exceeding one quarter, one half, and one full median target side are 27.3\%, 26.1\%, and 16.3\%, respectively. The unit-scale alignment burden is $\widehat{\mathcal B}_1=0.265$.}
    \label{fig:datasetstats}
\end{figure}

Among 300 sampled pairs, the estimated global shift has a mean of 4.66 pixels, median of 1.0 pixel, and standard deviation of 6.44 pixels. The integer estimate is zero for 46.0\% of the sample; nevertheless, 36.0\% exceed 2 pixels and 32.3\% exceed 5 pixels (Figure~\ref{fig:datasetstats}b). A zero estimated integer translation should not be interpreted as exact registration.

The finite search range censors the upper displacement tail. A search of $\pm12$ pixels along each axis has a maximum representable radial magnitude of

\begin{equation}
\sqrt{12^2+12^2}=16.97 \text{ pixels}.
\end{equation}

A pair assigned a boundary translation may have a larger true offset. Accordingly, 16.97 pixels is the maximum representable estimate, not a physical limit on camera misalignment.

Object-relative statistics are available for the 264 pairs with matching target annotations. Their median target side is 14.18 pixels; the mean and median alignment ratios are 0.297 and 0, respectively. The zero median reflects the prevalence of zero estimated integer shifts rather than a claim of perfect registration.

The burden fractions reveal a substantial upper tail: 27.3\% of annotation-matched pairs exceed one quarter of a median target side, 26.1\% exceed one half, and 16.3\% exceed a full side. Thus, roughly one in six measured pairs has an estimated global translation larger than the median annotated target (Figure~\ref{fig:datasetstats}c).

Integrating the empirical burden curve over the first target-side unit yields

\begin{equation}
\widehat{\mathcal B}_1=0.265.
\end{equation}

The value 0.265 summarizes the clipped target-relative shift (Eq.~\ref{eq:burdenscore}); the three threshold fractions retain information about how frequently the larger displacements arise.

Taken together, the target-size and shift distributions show why pixel displacement alone is insufficient to characterize the paired observations. The object-relative measures connect the estimated translation to the spatial extent of the objects under study.

These measurements concern global integer translations and not exact target correspondences. They do not account for parallax, lens distortion, depth variation, or spatially varying shifts. Their purpose is a reproducible geometric characterization of the sampled paired observations.

\section{AVID-C: a controlled perturbation benchmark}
\label{sec:avidc}

\subsection{Perturbation families and severity levels}

In addition to geometric mismatch, small-target recognition is sensitive to changes in image formation. Noise affects contrast, blur suppresses fine structure, and illumination variation or camera motion can alter the target and its surrounding background. Because RGB and infrared sensors respond differently, the affected modality is itself an important experimental variable.

\corrupt{} defines seven simulated degradation families: Gaussian noise, speckle noise, salt-and-pepper noise, Gaussian blur, uneven illumination, motion blur, and camera instability. Each family has five severity levels, generated from clean \dataset{} frames using specified parameters. The transformations can be reproduced without maintaining separate pre-rendered copies of every degraded condition.

Perturbation magnitudes are chosen for the native $320\times256$ resolution, where a blur kernel or translation of a few pixels can affect much of an 8--16 pixel target. The benchmark follows the controlled-corruption principle of ImageNet-C \citep{hendrycks2019benchmarking} without assuming that its image-size-dependent settings transfer unchanged to this regime. Table~\ref{tab:corruptions} specifies all five parameter settings per family.

\begin{table}[t]
\centering
\caption{\textbf{\corrupt{} perturbation families and severity levels.}
The table lists the parameter values used for severity levels 1--5. Noise standard deviations are expressed in 8-bit intensity units. Camera instability includes image translation, rotation, and a small variation in scale.}
\label{tab:corruptions}
\small
\begin{tabularx}{\linewidth}{@{}l l X@{}}
\toprule
Family & Parameter & Severity levels 1--5 \\
\midrule

Gaussian noise
& standard deviation
& 5, 10, 18, 28, 40 \\

Speckle noise
& multiplicative $\sigma$
& 0.05, 0.10, 0.18, 0.28, 0.40 \\

Salt-and-pepper
& corrupted pixel fraction
& 0.005, 0.01, 0.02, 0.04, 0.08 \\

Gaussian blur
& Gaussian $\sigma$ (px)
& 0.5, 1.0, 2.0, 3.5, 5.5 \\

Uneven illumination
& field strength
& 0.10, 0.20, 0.35, 0.50, 0.70 \\

Motion blur
& kernel length (px)
& 3, 7, 11, 17, 25 \\

Camera instability
& maximum shift / rotation
& (1 px, $0.5^\circ$), (3, $1.5^\circ$), (6, $3^\circ$),
  (10, $5^\circ$), (15, $8^\circ$) \\

\bottomrule
\end{tabularx}
\end{table}

Each perturbation is evaluated in three sensor configurations: RGB only, infrared only, and both modalities. Holding one stream clean isolates changes to the other; degrading both tests the combined observation under joint loss of image quality.

Results should be stratified by perturbation family, severity, and affected modality. Such reporting preserves sensor-specific failure patterns that a single corruption average would hide.

\subsection{Relating the perturbation scale to clean imagery}

The lowest Gaussian-noise severity is compared with the estimated noise already present in the unmodified imagery. This provides an image-level reference for the injected standard deviation, without treating a no-reference estimate as a physical sensor calibration.

Noise is estimated from 200 frames per modality using Immerkaer's no-reference Laplacian estimator \citep{immerkaer1996}. Because high-frequency scene structure can influence this estimate, the reported values characterize image content approximately rather than sensor electronics directly.

On the 0--255 intensity scale, the median full-frame estimate is 0.302 for RGB and 0.638 for infrared. Gaussian noise at severity 1 has an injected standard deviation of 5 intensity units, corresponding to approximately 16.6 and 7.8 times those estimates, respectively.

When estimation is restricted to the upper third of each frame, where sky is often less textured, the medians fall to 0.027 for RGB and 0.047 for infrared. This difference is consistent with the estimator's sensitivity to image structure.

The comparison in Figure~\ref{fig:corruptionanalysis}i confirms that the first injected Gaussian-noise level is above the estimated clean-frame noise for both modalities. It supports the chosen starting magnitude as a distinct image perturbation, but does not establish equivalence to noise produced by a particular physical sensor.

\section{A fixed-representation analysis of the dataset}
\label{sec:intrinsic}

\subsection{Measuring class structure under controlled variation}

The preceding measurements establish the scale and geometric properties of \dataset{}. To examine their consequences for bird-drone discrimination independently of a particular detector, we compare class-conditional distributions in a fixed patch representation. Target size, simulated degradation, and added cross-modal shift are varied while the representation remains unchanged.

Each annotated target defines a square crop centered on its bounding box, with $1.5\times$ padding. RGB and infrared crops are converted to grayscale, resized to $24\times24$, and flattened. Separate PCA bases fitted on clean patches reduce each modality to eight components. Their concatenation gives a 16-dimensional paired feature, compared with eight dimensions for each single-sensor feature.

\begin{figure}[t]
    \centering
    \includegraphics[width=0.99\linewidth]{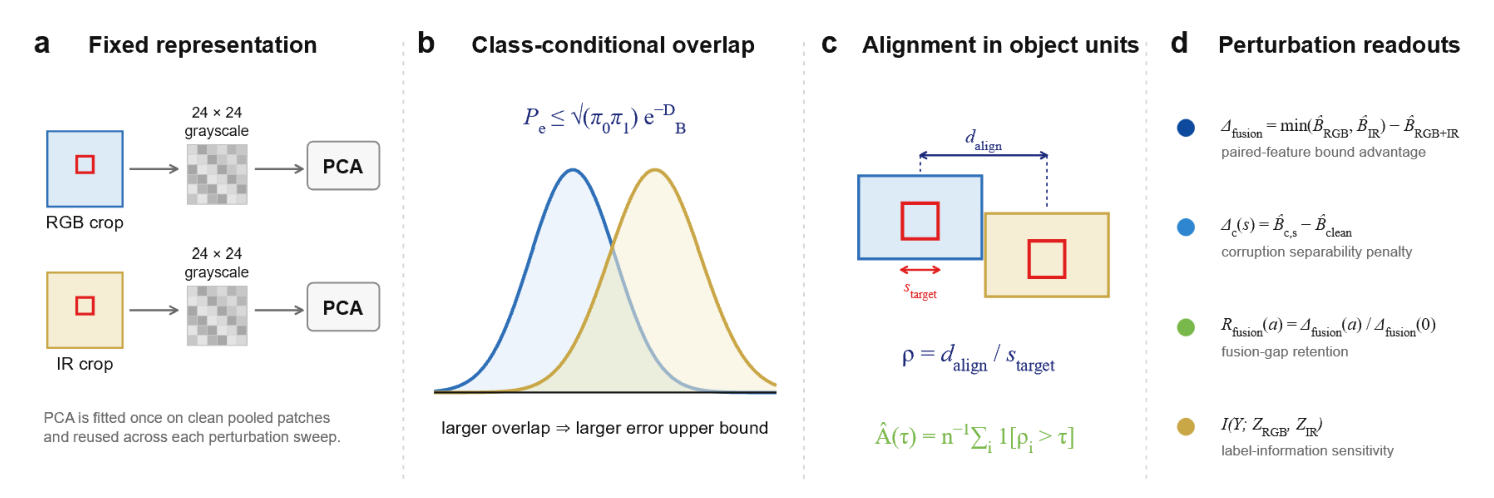}
    \caption{\textbf{Overview of the fixed-representation analysis used to study \dataset{}.}
    \textbf{a}, Annotated RGB and infrared target patches are converted to $24\times24$ grayscale images and projected into PCA spaces learned from the clean data.
    \textbf{b}, Bird and drone feature distributions are compared using Bhattacharyya and Chernoff error bounds, together with a kernel-density estimate as a non-Gaussian comparison.
    \textbf{c}, Cross-modal displacement is measured relative to target size using $\rho=d_{\mathrm{align}}/s_{\mathrm{target}}$.
    \textbf{d}, The resulting measurements are used to compare RGB, infrared, and combined features and to examine the effects of image degradation and increasing sensor misalignment.}
    \label{fig:intrinsicframework}
\end{figure}

The clean-data PCA bases are held fixed across all conditions, so changes in class overlap cannot be attributed to refitting the projection for each perturbation. Figure~\ref{fig:intrinsicframework} summarizes the feature extraction and analysis.

\subsection{Class overlap in the fixed representation}

Class structure is measured through the overlap of the bird and drone feature distributions. Greater overlap corresponds to weaker separation in the specified representation, without requiring a trained detector or an assumed full-frame localization mechanism.

Let $Z$ denote the feature vector and $Y\in\{0,1\}$ the bird or drone label. Each conditional feature distribution is approximated by a Gaussian,

\begin{equation}
Z \mid Y=y \sim \mathcal{N}(\mu_y,\Sigma_y).
\end{equation}

Here, $\mu_y$ and $\Sigma_y$ denote the class-specific mean and covariance estimates. The Bhattacharyya distance between the fitted Gaussian distributions is

\begin{align}
D_B
&=
\frac{1}{8}
(\mu_1-\mu_0)^\top
\bar{\Sigma}^{-1}
(\mu_1-\mu_0)
\\
&\quad+
\frac{1}{2}
\log
\frac{\det\bar{\Sigma}}
{\sqrt{\det\Sigma_0\det\Sigma_1}},
\qquad
\bar{\Sigma}
=
\frac{\Sigma_0+\Sigma_1}{2}.
\end{align}

The corresponding upper bound on classification error under class priors $\pi_0$ and $\pi_1$ is

\begin{equation}
P_e
\leq
\sqrt{\pi_0\pi_1}\,
e^{-D_B}.
\label{eq:bhatt}
\end{equation}

With equal bird and drone priors, we denote this fitted bound by $\Bhat$. Lower values indicate better class separation under the fitted Gaussian model. Because the target location and feature representation are supplied, $\Bhat$ is not an observed detector error rate and cannot establish full-frame recognition performance.

Two complementary estimates test the dependence on the density model. An optimized Chernoff calculation minimizes the bound over its weighting parameter, while a kernel-density estimate (KDE) provides a non-Gaussian overlap comparison. We report these cross-checks at the dominant 12--16 pixel target scale.

\subsection{Comparing single and paired modalities}

To compare paired features with the better individual modality, we define the fitted fusion gap as

\begin{equation}
\bfg
=
\min\!\left(
\Bhat_{\rgb},
\Bhat_{\ir}
\right)
-
\Bhat_{\rgb+\ir}.
\label{eq:bfg}
\end{equation}

Here $\Bhat_{\rgb}$, $\Bhat_{\ir}$, and $\Bhat_{\rgb+\ir}$ refer to RGB, infrared, and concatenated features. Positive values indicate a lower fitted bound for the paired representation, zero indicates no difference from the better single modality, and negative values favor that single modality.

The gap is a within-representation diagnostic. In particular, the principal comparison uses 16 paired components but only eight per single sensor; the magnitude of the gap therefore cannot be attributed solely to cross-sensor complementarity without accounting for feature dimension.

Equal-total-dimension and redundant-feature controls are evaluated in Section~\ref{sec:findings} to assess this sensitivity. Neither a positive fusion gap nor a smaller fitted bound is interpreted as a guaranteed increase in detector accuracy.

\subsection{Effect of corruption}

For corruption family $c$ and severity $s$, the fitted corruption penalty measures the change in the error bound relative to the clean condition:

\begin{equation}
\Delta_c(s)
=
\Bhat_{c,s}
-
\Bhat_{\mathrm{clean}}.
\label{eq:corruptpenalty}
\end{equation}

Here $\Bhat_{c,s}$ denotes the degraded-condition bound and $\Bhat_{\mathrm{clean}}$ its clean baseline. Positive penalties indicate increased fitted overlap, while negative penalties indicate increased fitted separation. Neither sign alone identifies the physical mechanism responsible.

A non-monotonic response is possible: mild smoothing, for example, may suppress nuisance variation before stronger blur removes discriminative target structure. The measured penalty characterizes the representation rather than presuming monotonic task degradation.

Separate clean baselines are used for RGB and infrared, so reported penalties reflect changes within each modality instead of differences between the underlying evaluation subsets.

\subsection{Sensitivity to cross-modal shift}

Controlled alignment stress tests displace infrared target patches relative to unchanged RGB patches. These shifts supplement, rather than replace, the estimates of native cross-modal displacement in Section~\ref{sec:alignment}.

Let $a$ be the maximum horizontal or vertical offset permitted in a trial. The persistence of the paired-feature advantage is summarized by fusion-gap retention

\begin{equation}
R_{\mathrm{fusion}}(a)
=
\frac{\bfg(a)}
{\bfg(0)}.
\label{eq:retention}
\end{equation}

Here $\bfg(a)$ is the fusion gap at shift budget $a$ and $\bfg(0)$ is the unshifted reference. A value of one indicates unchanged gap; values below one indicate its reduction. The ratio is interpreted only when the reference gap is nonzero.

The four diagnostics answer distinct questions: fitted bounds compare class overlap, the fusion gap compares feature configurations, corruption penalties compare image-quality conditions, and retention measures the response to added spatial disagreement. All share the same PCA representation, allowing controlled comparisons without implying detector-level performance.

The accompanying analysis reports both the main estimates and representation-dimension checks, so numerical trends can be considered alongside their modeling assumptions.

\section{Dataset behavior under scale, corruption, and alignment}
\label{sec:findings}

\subsection{Target scale shapes class separation}

\begin{figure}[t]
    \centering
    \includegraphics[width=0.99\linewidth]{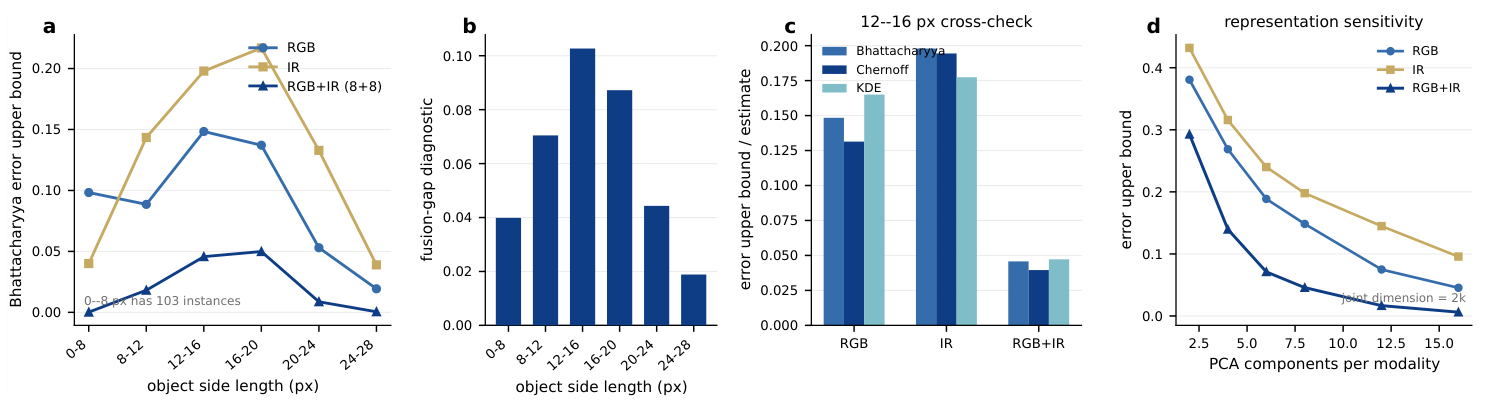}
    \caption{\textbf{Effect of target size on bird-drone class separation.}
    \textbf{a}, Fitted Bhattacharyya error bounds across target-size intervals for RGB, infrared, and the combined representation using eight PCA components from each modality. The 0--8 pixel interval contains only 103 annotated instances and should be interpreted with caution.
    \textbf{b}, Fusion gap calculated using Eq.~\ref{eq:bfg} for different target sizes.
    \textbf{c}, Comparison of Bhattacharyya bounds, optimized Chernoff bounds, and KDE estimates for targets between 12 and 16 pixels.
    \textbf{d}, Effect of the number of PCA components on the fitted error bounds. The combined representation uses $k$ components from each modality, giving a total of $2k$ features.}
    \label{fig:scaleanalysis}
\end{figure}

Figure~\ref{fig:scaleanalysis}a presents the fitted class-overlap bounds across native target-size intervals. Scale changes the available visual evidence, and the comparison shows whether paired features retain separation when the objects are represented by very few pixels.

The 8+8-dimensional paired representation has the lowest fitted bound in every size interval with sufficient examples from both classes. For the dominant 12--16 pixel interval, the Bhattacharyya values are $\Bhat_{\rgb}=0.148$, $\Bhat_{\ir}=0.198$, and $\Bhat_{\rgb+\ir}=0.046$, giving a fusion gap of 0.103 (Figure~\ref{fig:scaleanalysis}b). The ordering indicates greater fitted separation in the combined feature space at this scale, but does not isolate a causal benefit from the increased number of components.

Alternative density comparisons support the same ordering at 12--16 pixels. Optimized Chernoff bounds are 0.131 for RGB, 0.194 for infrared, and 0.040 for paired features; KDE estimates are 0.165, 0.177, and 0.047 (Figure~\ref{fig:scaleanalysis}c). Their agreement in rank suggests that the paired ordering is not specific to the Bhattacharyya formula, although all results remain conditional on the extracted features.

Feature dimensionality is a substantive confound. The main paired vector has 16 components, whereas each single-sensor vector has eight; as shown in Figure~\ref{fig:scaleanalysis}d, fitted bounds generally decline as the number of retained PCA components increases.

At equal total dimension, four RGB and four infrared components yield a paired bound of 0.140, compared with 0.148 for eight RGB components and 0.198 for eight infrared components. The paired bound remains lower at 12--16 pixels, but its margin is substantially reduced, and the ordering does not extend uniformly to all larger-scale intervals.

Redundant-dimension controls further reveal sensitivity in the fitted estimator. Duplicating an eight-component RGB vector changes its bound from 0.148 to 0.127; duplicating infrared changes 0.198 to 0.191. Since duplicated features contain no new information, the changes are consistent with the influence of covariance estimation or regularization. The 16-dimensional paired bound of 0.046 remains smaller, but its gap must be interpreted as a property of the chosen fitted model, not as a dimension-independent measure of sensor complementarity.

\subsection{RGB and infrared respond differently to corruption}

\begin{figure}[t]
    \centering
    \includegraphics[width=0.92\linewidth]{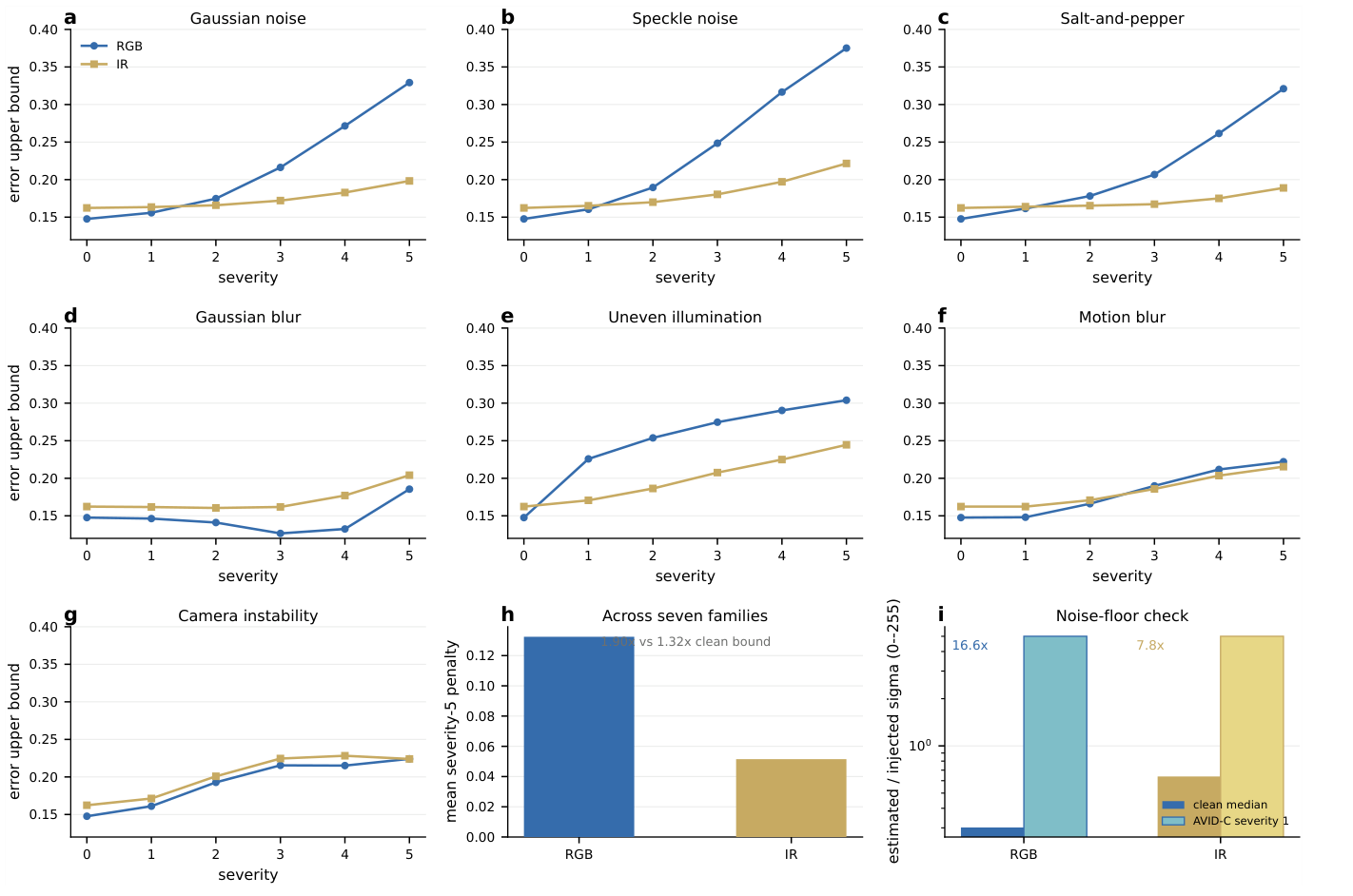}
    \caption{\textbf{Effect of simulated image degradation on RGB and infrared class separation.}
    Panels \textbf{a--g} show the fitted Bhattacharyya error bounds from clean images (severity 0) through severity level 5 for each of the seven perturbation families.
    \textbf{h}, Mean increase in the fitted error bound at severity level 5 relative to the corresponding clean baseline.
    \textbf{i}, Estimated noise levels in clean RGB and infrared images compared with Gaussian-noise severity level 1. The added noise has a standard deviation approximately 16.6 times the median RGB estimate and 7.8 times the median infrared estimate.}
    \label{fig:corruptionanalysis}
\end{figure}

The \corrupt{} analysis uses the same 1,500 target instances for every perturbation and severity, with PCA bases fixed from clean images. Figure~\ref{fig:corruptionanalysis}a--g reports fitted bounds across severity levels 0--5 for each family and modality.

Clean bounds for this evaluation subset are 0.148 for RGB and 0.162 for infrared. These differ from the 12--16 pixel scale-analysis baselines because the sample compositions are different; the two experiments should not be compared without accounting for that distinction.

At severity 5, RGB bounds are 0.329 for Gaussian noise, 0.375 for speckle noise, 0.321 for salt-and-pepper noise, 0.186 for Gaussian blur, 0.304 for uneven illumination, 0.222 for motion blur, and 0.224 for camera instability. Infrared bounds for the same families are 0.198, 0.222, 0.189, 0.204, 0.245, 0.215, and 0.224. The changes are therefore strongly dependent on both degradation type and modality.

Noise perturbations produce the clearest modality asymmetry. At severity 5, speckle noise increases the RGB bound from 0.148 to 0.375, compared with 0.162 to 0.222 for infrared. Gaussian and salt-and-pepper noise show a similar ordering, suggesting greater retention of fitted infrared class separation under these particular simulated noise models.

Other perturbations show smaller differences. Gaussian-blur bounds are 0.186 for RGB and 0.204 for infrared; motion-blur bounds are 0.222 and 0.215. Camera instability yields approximately 0.224 for both. These results do not support a blanket claim that one modality is more robust across all degradation families.

Averaged over the seven severity-5 conditions, the clean-relative increase is 0.132 for RGB and 0.052 for infrared (Figure~\ref{fig:corruptionanalysis}h). Mean severity-5 bounds are 1.90 and 1.32 times their respective clean baselines. The family-level results remain essential: the averages summarize the selected transformations, not a universal ranking of sensor robustness.

Gaussian blur illustrates the non-monotonic nature of fitted overlap: the RGB bound initially decreases and rises at stronger severity. Mild smoothing could suppress irrelevant high-frequency variation before larger kernels remove target structure, although the experiment does not establish that mechanism. Nominal perturbation severity should therefore not be equated directly with a monotonic change in class separability.

\subsection{Cross-modal shift weakens paired separation}

The alignment stress test shifts infrared patches while holding RGB fixed. Horizontal and vertical translations are sampled independently within each per-axis budget; five trials are evaluated for each nonzero budget. The feature representation is unchanged.

Table~\ref{tab:alignstressmain} reports budgets through 12 pixels per axis, consistent with the native-image NMI search range. The budget is an interval bound on individual offsets, not an imposed radial shift of that magnitude.

The fusion gap decreases from 0.111 without added shift to 0.093 at a 12-pixel budget, yielding

\begin{equation}
R_{\mathrm{fusion}}(12)=0.844,
\end{equation}

Thus 84.4\% of the original fitted gap remains at the largest budget in the main comparison.

\begin{table}[t]
\centering
\caption{\textbf{Effect of additional infrared shift on fitted class separation.}
The shift budget $a$ gives the maximum allowed displacement along each image axis, rather than a fixed radial displacement. Each nonzero condition is repeated over five trials.}
\label{tab:alignstressmain}
\small
\begin{tabular}{rrrr}
\toprule
Shift budget $a$ (px) & Paired bound & Fusion gap & Gap retention \\
\midrule
0  & 0.0370 & 0.1107 & 1.000 \\
1  & 0.0383 & 0.1094 & 0.988 \\
2  & 0.0414 & 0.1063 & 0.960 \\
5  & 0.0472 & 0.1005 & 0.907 \\
8  & 0.0507 & 0.0970 & 0.876 \\
12 & 0.0542 & 0.0935 & 0.844 \\
\bottomrule
\end{tabular}
\end{table}

Over the same sweep, the paired bound rises from 0.0370 to 0.0542. Additional displacement therefore weakens the fitted paired-feature separation in this experiment, although the magnitude is specific to the feature representation and should not be interpreted as a detector accuracy loss.

Native-shift measurements and controlled shifts address different questions. Figure~\ref{fig:datasetstats}c quantifies translation estimates present in observed frame pairs; Table~\ref{tab:alignstressmain} measures the response of the fixed feature representation to additional displacement. Their numerical values are not interchangeable.

\subsection{Scale, corruption, and alignment describe different sources of difficulty}

Target scale, image degradation, and inter-sensor displacement represent different sources of difficulty. Scale limits the number of pixels containing object evidence; degradation alters that evidence; and displacement changes the spatial relation between the two observations.

Their interaction is especially consequential for \dataset{} because most targets have sides between 8 and 16 pixels. In this regime, a few pixels of blur or translation can affect a large proportion of the target region. The analyses show both modality-specific corruption responses and reduced fitted paired-feature separation under added shift.

Accordingly, aggregate detection or classification results should be accompanied by target-scale, perturbation, and alignment strata. Such measurements provide information about operating conditions that a single overall performance score cannot convey.

\section{Benchmark use and reporting}
\label{sec:protocol}

\dataset{} and \corrupt{} are intended for evaluations in which the acquisition and perturbation conditions are explicit. Because the corpus is dominated by tiny targets and the two modalities may differ geometrically and radiometrically, task metrics are most informative when stratified by target size, sensing modality, degradation, and alignment. Table~\ref{tab:reporting} summarizes the recommended protocol.

\subsection{Clean paired evaluation}

Clean-data evaluations should specify the input modalities (RGB, infrared, or both), source videos, and any cropping or resizing relative to the native $320\times256$ resolution. Training and evaluation partitions should be disjoint at the video level: splitting adjacent frames across sets risks information leakage. The original Test partition is not part of the paired RGB-infrared benchmark.

\subsection{Scale-aware evaluation}

For object-level reporting, the geometric-mean side length
\begin{equation}
s=\sqrt{w_{\mathrm{px}}h_{\mathrm{px}}}
\end{equation}
should be computed from the native annotated boxes. We recommend intervals of 0--8, 8--12, 12--16, 16--20, 20--24, 24--28, 28--32, and 32--48 pixels, with instance counts accompanying each score. The 0--8 pixel group contains only 103 annotations and warrants particular caution. For sequence-level tasks, the distribution or median of native target sizes can characterize the scale regime of each clip.

\subsection{Corruption evaluation}

\corrupt{} results should specify the degradation family, severity level, and affected modality. The three standard configurations are RGB-only, infrared-only, and joint degradation. For joint perturbations, the relationship between random realizations across streams should also be stated. Severity curves and family-specific results are preferable to an isolated average because perturbation responses can be modality-dependent and non-monotonic.

\subsection{Alignment-aware evaluation}

Alignment reports should state the estimator, search range, and displacement units. In addition to pixel magnitude $d_{\mathrm{align}}$, we recommend the object-relative ratio
\begin{equation}
\rho=\frac{d_{\mathrm{align}}}{s_{\mathrm{target}}}
\end{equation}
and burden fractions $\burden(0.25)$, $\burden(0.5)$, and $\burden(1.0)$, together with $\widehat{\mathcal B}_1$ when a scalar summary is useful. These thresholds correspond to shifts exceeding one quarter, one half, and one median target side. Results obtained with different registration estimators should not be compared as though their displacement estimates were interchangeable; the target-size definition must also be consistent.

\subsection{Comparison with the fixed-representation analysis}

Numerical comparison with the class-overlap bounds in Section~\ref{sec:intrinsic} requires matched target crops, $24\times24$ grayscale inputs, PCA dimensions, class priors, and density-fitting procedures. Alternative representations are valuable for studying generality, but their absolute bounds and fusion gaps should be reported as separate experimental settings rather than directly equated with the present estimates.

\begin{table}[t]
\centering
\caption{\textbf{Recommended reporting details for experiments using \dataset{} and \corrupt{}.}
Reporting the evaluated data, target sizes, sensor conditions, and experimental settings helps make results easier to reproduce and compare.}
\label{tab:reporting}
\small
\begin{tabularx}{\linewidth}{@{}l X@{}}
\toprule
Item & What to report \\
\midrule

Input
& RGB, infrared, or both modalities; original or resized image resolution \\

Data subset
& Dataset split, video identifiers when relevant, and number of frames \\

Target scale
& Geometric-mean bounding-box size intervals and number of instances \\

Corruption
& Perturbation type, severity level, and affected modality or modalities \\

Alignment
& Estimation method, search range, and object-relative alignment thresholds or burden curve \\

Aggregation
& Results across target sizes and perturbation types, alongside any overall average \\

Fixed representation
& Crop definition, PCA dimensions, class priors, and density model for comparisons with Section~\ref{sec:intrinsic} \\

\bottomrule
\end{tabularx}
\end{table}

\section{Data access and licensing}
\label{sec:repro}

The \dataset{} corpus is hosted at \url{https://huggingface.co/datasets/Dharini24/AVID}, with corresponding RGB-infrared frames, existing annotations, and source video identifiers. The source recordings were published in the Drone Detection Dataset under CC0 1.0 Universal \citep{svanstrom2020repository}; the original authors request citation of the associated dataset publication \citep{svanstrom2021multisensor}. The AVID dataset package is listed as CC BY-NC 4.0, which does not revoke the original CC0 status of source imagery. The alignment analysis, \corrupt{} implementation, and dataset preparation code are available at \url{https://github.com/rdharini2001/ICIP_2025_Data_Release} under the MIT license. Users should identify the dataset version and cite both the source collection and this benchmark.

\section{Responsible use}
\label{sec:responsible}

\dataset{} contains distant surveillance observations of birds and drones; it has no person-identity or personal-attribute labels. The benchmark can support research on aerial-target perception and sensor robustness, including the distinction between birds and drones that might otherwise be conflated with background. Counter-drone surveillance nevertheless raises deployment risks, including intrusive monitoring and consequential false alarms. Performance on these data is not evidence of operational safety or legal compliance. Deployment requires independent assessment of environmental transfer, false positives, human oversight, and the consequences of incorrect decisions.

\section{Limitations}
\label{sec:limitations}

\paragraph{Corpus scope and annotation provenance.}
The paired benchmark excludes the original VIP Cup Test export because its RGB and infrared frame identities cannot be matched unambiguously using the available information. The dataset retains annotations prepared for the earlier challenge; a complete record of annotation assignment and adjudication is not available, so agreement and label uncertainty cannot be quantified independently.

\paragraph{Alignment sampling and model.}
Translation statistics are estimated for 300 sampled pairs rather than the entire corpus, and only 264 have matching annotations for target-relative normalization. NMI estimates one global integer translation per frame pair, not local correspondence. Parallax, lens distortion, differences in viewpoint, and target-dependent offsets are not resolved. The $\pm12$-pixel per-axis search also censors the largest possible displacements; 16.97 pixels is the maximum representable radial estimate, not a physical upper bound.

\paragraph{Representation dependence.}
The fixed-feature analysis assumes known target locations and uses grayscale crops, PCA, and fitted class densities. It measures neither full-frame localization nor temporal tracking. Paired vectors contain more components than single-modality vectors, and redundant-feature experiments reveal numerical sensitivity to dimension and covariance treatment. Equal-dimension comparisons reduce the fitted fusion gap. KDE relaxes the Gaussian density assumption but does not remove dependence on the features themselves. None of these estimates is measured detector accuracy.

\paragraph{Simulated degradation and generalization.}
\corrupt{} covers seven selected perturbation families, not the full diversity of atmospheric effects, thermal artifacts, compression, lens contamination, or complex camera motion. Increasing simulated severity need not monotonically increase fitted overlap or downstream task error. Finally, the 115 source videos cannot represent all geographic regions, weather, sensor models, airframes, bird species, or observation distances. Claims about deployment or broad generalization require independent datasets and task-level evaluations.

\section{Conclusion}

\dataset{} provides a focused benchmark for a common but difficult surveillance setting: birds and drones represented by only a few pixels in corresponding, imperfectly aligned RGB and infrared frames. Built from the Drone Detection Dataset imagery and the IEEE ICIP 2025 VIP Cup bird-drone curation \citep{svanstrom2021multisensor}, it contains 32,040 frame pairs from 115 videos and 35,952 existing annotations. Its analysis makes the operating conditions measurable. In particular, 68.1\% of targets have sides of 8--16 pixels, and 16.3\% of 264 annotation-matched pairs have an estimated shift greater than one median target side.

The \corrupt{} benchmark extends this characterization to sensor-specific simulated degradation. Fixed-feature experiments show that RGB and infrared respond differently to several noise conditions and that the fitted paired-modality advantage decreases under additional spatial shift. At the dominant 12--16 pixel scale, combined features show lower fitted class-overlap bounds, but dimension controls limit how broadly that difference can be interpreted. These findings concern the stated representation, not trained detector performance. By connecting scale, alignment, and image quality to evaluation conditions, \dataset{} and \corrupt{} support more informative comparisons of methods for tiny aerial-target perception.

\section*{Acknowledgements}

We thank Fredrik Svanstr\"om, Fernando Alonso-Fernandez, and Cristofer Englund for collecting and sharing the original Drone Detection Dataset \citep{svanstrom2021multisensor}. The bird-drone material was curated for the IEEE ICIP 2025 Video and Image Processing Cup, ``Infrared-Visual Fusion for Enhanced Drone Detection, Tracking and Payload Identification in Surveillance Videos.'' We acknowledge the wider VIP Cup team for its contributions to the challenge. The present study develops the paired \dataset{} benchmark, target-relative alignment measures, simulated \corrupt{} perturbations, and accompanying analysis from that earlier curation.

\bibliographystyle{unsrtnat}
\bibliography{references}

\clearpage
\appendix

\section{Dataset documentation}
\label{app:datasheet}

The following dataset documentation follows the themes of Datasheets for Datasets \citep{gebru2021datasheets}, with additional information on provenance, composition, annotation scope, and maintenance.

\subsection{Motivation}

\textbf{Where does the imagery come from?}

The source imagery was collected by Svanstr\"om et al. for the Drone Detection Dataset \citep{svanstrom2021multisensor}. The bird-drone portion was curated for the IEEE ICIP 2025 VIP Cup \citep{ieeevipcup2025}; no new video capture is claimed in this study.

\textbf{What is new in \dataset{} and \corrupt{}?}

\dataset{} comprises corresponding bird-drone RGB-infrared observations with inherited bounding boxes and video identifiers. The new benchmark components are target-relative alignment measures, \corrupt{} simulations, and fixed-feature class-separation analyses. Payload identification is outside this study's scope.

\subsection{Composition}

\textbf{What does the paired dataset contain?}

\dataset{} contains 32,040 corresponding RGB-infrared frame pairs from 115 surveillance videos, with 35,952 annotated bird and drone instances. The Train split includes 28,790 frame pairs from 102 videos, while the Validation split contains 3,250 pairs from 13 videos. The annotations consist of 15,471 bird instances and 20,481 drone instances.

Pair identity denotes corresponding observations of a scene, not registration at the pixel level.

\textbf{What is the image resolution?}

Both RGB and infrared frames have a native resolution of $320\times256$ pixels. The dataset statistics and target-size measurements reported in this paper refer to this original resolution.

\textbf{How long are the video sequences?}

The 115 video sequences contain between 50 and 329 paired frames. The median sequence length is 306 frames, and the mean is 278.6 frames.

\textbf{How small are the targets?}

Target sizes are strongly concentrated in the small-object range: 68.1\% have a geometric-mean box side of 8--16 pixels and 99.0\% satisfy the COCO small criterion. The fine-grained counts are shown in Table~\ref{tab:scalehist}.

\begin{table}[h]
\centering
\caption{\textbf{Distribution of target sizes in \dataset{}.}
Target size is measured using the geometric-mean side length of each annotated bounding box.}
\label{tab:scalehist}
\small
\begin{tabular}{lrrrrrrrrr}
\toprule
Side (px)
& 0--4 & 4--8 & 8--12 & 12--16 & 16--20
& 20--24 & 24--28 & 28--32 & 32--48 \\
\midrule
Count
& 0 & 103 & 6,783 & 17,706 & 7,269
& 2,155 & 1,189 & 376 & 371 \\
\bottomrule
\end{tabular}
\end{table}

\textbf{What is not included?}

The benchmark concerns bird-drone surveillance and does not include the payload-identification task. The VIP Cup Test split is also outside the paired evaluation corpus because reliable RGB-infrared correspondence is unavailable for that partition.

\subsection{Collection and annotation}

\textbf{What viewpoint does the dataset represent?}

The camera viewpoint is ground-based and directed toward distant airborne birds and drones \citep{svanstrom2021multisensor}, not a UAV-mounted camera viewing ground objects.

\textbf{Where do the bounding boxes come from?}

Bounding boxes and bird-drone class assignments originate in the VIP Cup curation. The available records do not fully document annotation review or adjudication, preventing independent estimation of annotator agreement.

\subsection{Access and maintenance}

\textbf{Where is the dataset maintained?}

The paired dataset is hosted at \url{https://huggingface.co/datasets/Dharini24/AVID}. Benchmark implementations, alignment measures, and scripts are available at \url{https://github.com/rdharini2001/ICIP_2025_Data_Release}.

\textbf{What license applies to the data and code?}

The original source imagery was released under CC0 1.0 Universal \citep{svanstrom2020repository}. The \dataset{} listing specifies CC BY-NC 4.0 for the released dataset package, but this does not change the original CC0 terms that apply to the source imagery. The project code is released under the MIT license.

\section{Alignment measurement}
\label{app:alignment}

Global RGB-infrared translation is estimated for a fixed random sample of 300 frame pairs from the Train and Validation corpus. The procedure searches integer displacement and measures relative shift for annotation-matched pairs.

RGB frames are converted to grayscale, and infrared frames are brought to matching spatial dimensions where required. Candidate horizontal and vertical translations are considered over

For each frame pair, we evaluate integer translations along the horizontal and vertical axes within the range

\begin{equation}
(d_x,d_y)\in[-12,12]^2.
\end{equation}

For each candidate displacement, NMI is evaluated after excluding a 12-pixel border to avoid contamination by translated image boundaries.

NMI is computed from a 32-bin joint histogram as

\begin{equation}
\mathrm{NMI}(A,B)
=
\frac{H(A)+H(B)}
{H(A,B)},
\end{equation}

where $H(A)$ and $H(B)$ are the individual image entropies and $H(A,B)$ is their joint entropy.

The translation maximizing NMI defines the estimated global shift. Its Euclidean magnitude is

\begin{equation}
d_{\mathrm{align}}
=
\sqrt{d_x^2+d_y^2}.
\end{equation}

The sample is fixed for reproducibility, but its translation distribution need not match that of all 32,040 frame pairs.

A Sobel-edge phase-correlation estimate provides a secondary cross-check. Neither the NMI nor phase-correlation estimate is registration ground truth; low texture and cross-modal intensity differences can affect both.

The displacement is normalized by the median geometric-mean annotated target side $s_i$ within frame pair $i$, giving

\begin{equation}
\rho_i=\frac{d_i}{s_i}.
\end{equation}

A ratio of 0.5 denotes an estimated shift equal to half the median target side. The empirical burden curve records the fraction exceeding each threshold:

We summarize the frequency of different relative displacements using the empirical alignment burden curve:

\begin{equation}
\widehat A(\tau)
=
\frac{1}{n}
\sum_{i=1}^{n}
\mathbf{1}[\rho_i>\tau].
\end{equation}

Here $n$ counts annotation-matched frame pairs and $\tau$ is the relative-shift threshold.

The unit-scale truncated burden is

\begin{equation}
\widehat{\mathcal B}_{1}
=
\int_0^1
\widehat A(\tau)\,d\tau
=
\frac{1}{n}
\sum_i
\min(\rho_i,1).
\end{equation}

The score is the mean ratio after clipping values above one. The threshold curve retains information about the distribution that the scalar alone cannot convey.

Table~\ref{tab:alignsummary} distinguishes the 300 raw-shift estimates from the 264 pairs that support target-relative normalization.

Across all 300 pairs, the mean estimated shift is 4.66 pixels (median 1.00; standard deviation 6.44). Of these, 46.0\% have a zero estimated integer translation and 32.3\% exceed 5 pixels.

For the 264 annotation-matched pairs, the median target side is 14.18 pixels. Burden fractions at $\tau=0.25,0.5,1$ are 27.3\%, 26.1\%, and 16.3\%, and the unit-scale burden is $\widehat{\mathcal B}_1=0.265$.

\begin{table}[h]
\centering
\caption{\textbf{Summary of RGB-infrared alignment measurements.}
Global translation is estimated for 300 frame pairs, of which 264 have matching target annotations for the object-relative analysis. The maximum representable shift is determined by the translation search range and does not indicate the largest possible physical displacement between the cameras.}
\label{tab:alignsummary}
\small
\begin{tabular}{lr}
\toprule
Quantity & Value \\
\midrule
Number of measured shifts & 300 \\
Pairs with matched annotations & 264 \\
Pairs without matched annotations & 36 \\
Mean estimated shift & 4.66 px \\
Median estimated shift & 1.00 px \\
Standard deviation & 6.44 px \\
Fraction at 0 px & 46.0\% \\
Fraction $>2$ px & 36.0\% \\
Fraction $>5$ px & 32.3\% \\
Maximum representable radial shift & 16.97 px \\
Median target side (264 pairs) & 14.18 px \\
$\widehat A(0.25)$ & 0.273 \\
$\widehat A(0.50)$ & 0.261 \\
$\widehat A(1.00)$ & 0.163 \\
$\widehat{\mathcal B}_{1}$ & 0.265 \\
\bottomrule
\end{tabular}
\end{table}

Because the per-axis search is limited to $\pm12$ pixels, translations with larger magnitudes can be clipped by the search boundary. The largest representable radial magnitude is 16.97 pixels.

The resulting statistics characterize global translations within this range, not local or target-specific pixel correspondence.

\section{Fixed-representation analysis details}
\label{app:intrinsic}

\subsection{Patch representation and scale analysis}

The analysis covers 35,952 annotated bird and drone instances. Each annotation yields a centered square crop with $1.5\times$ padding, converted to grayscale, resized to $24\times24$, and flattened.

Separate PCA projections are fitted on clean RGB and infrared patches. The principal representation retains eight components from each modality; concatenating them gives a 16-dimensional paired feature.

Class priors are set equal for all overlap estimates. The PCA bases are fixed across scale, corruption, and added-shift experiments.

Instances are stratified by native geometric-mean box side. A scale interval enters the fitted overlap comparison only if each class contributes at least 15 instances.

Bhattacharyya bounds for supported size intervals are listed in Table~\ref{tab:scalebounds}; lower bounds indicate greater fitted separation.

\begin{table}[h]
\centering
\caption{\textbf{Bhattacharyya error upper bounds across target sizes.}
The paired representation combines eight PCA components from RGB and eight from infrared. A dash indicates that the size interval does not meet the minimum sample requirement for both classes.}
\label{tab:scalebounds}
\small
\begin{tabular}{lrrrrrrr}
\toprule
& 0--8 & 8--12 & 12--16 & 16--20 & 20--24 & 24--28 & 28+ \\
\midrule
RGB
& 0.098 & 0.089 & 0.148 & 0.137 & 0.053 & 0.019 & - \\
Infrared
& 0.040 & 0.143 & 0.198 & 0.217 & 0.133 & 0.039 & - \\
RGB+IR (8+8)
& $<$0.001 & 0.018 & 0.046 & 0.050 & 0.009 & $<$0.001 & - \\
\bottomrule
\end{tabular}
\end{table}

The 0--8 pixel interval contains 52 birds and 51 drones (103 instances total). Although it meets the minimum class-count criterion, its estimates have less support than those from more populated bins.

\subsection{Chernoff and KDE comparison}

The Gaussian Bhattacharyya bound is supplemented by optimized Chernoff bounds and kernel-density estimation (KDE), allowing the class-overlap ordering to be assessed under different density assumptions.

The Bhattacharyya case fixes the Chernoff parameter at $s=0.5$. Optimization evaluates 99 values in the interval $[0.01,0.99]$ and retains the smallest resulting bound.

For the 12--16 pixel subset, KDE-based overlap is evaluated using 4,000 importance samples from an equal mixture of the fitted class densities. This comparison does not impose Gaussian class densities.

The 12--16 pixel interval is the most populated scale in \dataset{}; Table~\ref{tab:crosschecks} reports the three estimates for this regime.

\begin{table}[h]
\centering
\caption{\textbf{Comparison of class-overlap estimates for targets between 12 and 16 pixels.}
Results are reported for the fitted Bhattacharyya bound, optimized Chernoff bound, and KDE-based estimate.}
\label{tab:crosschecks}
\small
\begin{tabular}{lrrrr}
\toprule
Representation
& Bhattacharyya ($s=0.5$)
& Optimized Chernoff
& Optimal $s$
& KDE \\
\midrule
RGB
& 0.1484 & 0.1314 & 0.33 & 0.1650 \\
Infrared
& 0.1979 & 0.1944 & 0.43 & 0.1774 \\
RGB+IR
& 0.0457 & 0.0396 & 0.37 & 0.0472 \\
\bottomrule
\end{tabular}
\end{table}

All three calculations rank the combined RGB-infrared representation as having the lowest fitted overlap at 12--16 pixels.

The absolute values remain conditional on the PCA features and density estimators, and do not measure errors of a trained full-frame detector.

\subsection{Representation-dimension controls}

The main paired feature has twice the dimension of either individual modality. To test whether fitted separation partly reflects this difference, we include an equal-total-dimension comparison and redundant-feature controls.

The controls isolate two distinct aspects of dimensionality: substituting a second modality while holding total dimension fixed, and increasing dimension without adding information.

The equal-dimension paired representation retains four RGB and four infrared components, while each single-modality comparator retains eight.

The redundant controls repeat the eight-component RGB or infrared feature vector to produce 16 dimensions without any new sensor information.

Table~\ref{tab:dimcontrol} reports these comparisons in the dominant 12--16 pixel interval.

\begin{table}[h]
\centering
\caption{\textbf{Effect of feature dimension on fitted class separation at 12--16 pixels.}
The equal-dimension and duplicated-feature comparisons help identify how feature dimension influences the estimated bounds.}
\label{tab:dimcontrol}
\small
\begin{tabular}{lrrl}
\toprule
Representation & Total dimension & Bound & Composition \\
\midrule
RGB PCA
& 8 & 0.1484 & RGB only \\

Infrared PCA
& 8 & 0.1979 & infrared only \\

RGB+IR
& 8 & 0.1396 & 4 RGB + 4 IR \\

RGB duplicated
& 16 & 0.1272 & repeated RGB feature \\

Infrared duplicated
& 16 & 0.1913 & repeated infrared feature \\

RGB+IR
& 16 & 0.0457 & 8 RGB + 8 IR \\
\bottomrule
\end{tabular}
\end{table}

With eight total components, the paired bound is 0.1396, versus 0.1484 for RGB and 0.1979 for infrared. This is a smaller paired-versus-single difference than the principal 16-dimensional result.

Redundant duplication changes the fitted RGB bound from 0.1484 to 0.1272 and the infrared bound from 0.1979 to 0.1913. Because the data contain no additional information, this sensitivity points to covariance estimation, regularization, or other numerical aspects of fitting.

The controls justify interpreting the fusion gap within a specified feature dimension and fitting procedure rather than as a representation-independent measure of multimodal information.

\subsection{Corruption analysis}

The corruption experiments use a fixed subset of 1,500 annotated instances across all families and severities. PCA is fitted once to the clean data and retained for every degraded condition.

The clean Bhattacharyya baselines for this subset are 0.1477 in RGB and 0.1623 in infrared. Their difference from the scale-analysis baselines reflects the use of a different subset.

The severity-5 bounds for all seven perturbations are listed in Table~\ref{tab:severityfive}.

\begin{table}[h]
\centering
\caption{\textbf{Bhattacharyya error upper bounds at severity level 5 of \corrupt{}.}
The corresponding clean-image bounds are 0.1477 for RGB and 0.1623 for infrared.}
\label{tab:severityfive}
\small
\begin{tabular}{lrr}
\toprule
Perturbation & RGB & Infrared \\
\midrule
Gaussian noise & 0.3292 & 0.1982 \\
Speckle noise & 0.3752 & 0.2215 \\
Salt-and-pepper & 0.3211 & 0.1889 \\
Gaussian blur & 0.1855 & 0.2041 \\
Uneven illumination & 0.3040 & 0.2446 \\
Motion blur & 0.2221 & 0.2155 \\
Camera instability & 0.2241 & 0.2240 \\
\bottomrule
\end{tabular}
\end{table}

The largest modality differences occur under Gaussian, speckle, and salt-and-pepper noise, for which the fitted RGB bound rises more than its infrared counterpart.

Blur and camera-instability results show smaller between-modality differences, motivating separate reporting for each perturbation rather than inference from an aggregate score.

All values quantify fitted class separation in known target crops; none is a measured detector error rate.

\subsection{Controlled cross-modal shift}

The alignment stress test holds RGB target patches fixed and translates corresponding infrared patches to introduce controlled cross-modal disagreement.

The per-axis shift budget $a$ takes values

\begin{equation}
a\in\{1,2,5,8,12,17\}\text{ pixels}.
\end{equation}

For every nonzero budget, five trials sample horizontal and vertical translations independently from $[-a,a]$.

Displacements in native-image coordinates are mapped into the $24\times24$ patch space by a factor of $24/16=1.5$, chosen for the dominant 12--16 pixel target regime.

The principal results stop at a 12-pixel per-axis budget, matching the NMI search range; the 17-pixel stress condition is outside that comparison.

Table~\ref{tab:alignsweep} reports the corresponding paired bound, fusion gap, and retention.

\begin{table}[h]
\centering
\caption{\textbf{Effect of controlled infrared displacement on the paired feature representation.}
The shift budget specifies the maximum allowed displacement along each image axis rather than a fixed radial displacement. Nonzero conditions are evaluated over five trials.}
\label{tab:alignsweep}
\small
\begin{tabular}{rrrr}
\toprule
Shift budget
& Joint upper bound
& Fusion gap
& Gap retention \\
\midrule
0  & 0.0370 & 0.1107 & 1.000 \\
1  & 0.0383 & 0.1094 & 0.988 \\
2  & 0.0414 & 0.1063 & 0.960 \\
5  & 0.0472 & 0.1005 & 0.907 \\
8  & 0.0507 & 0.0970 & 0.876 \\
12 & 0.0542 & 0.0935 & 0.844 \\
\bottomrule
\end{tabular}
\end{table}

Across budgets 0--12 pixels, the paired bound increases from 0.0370 to 0.0542 and the fusion gap decreases from 0.1107 to 0.0935, leaving 84.4\% of the initial gap. This is a feature-space stress test, not a prediction of tracking or detection accuracy.

The trend demonstrates sensitivity to added displacement under the selected fixed representation. It cannot establish the sensitivity of a trained detector to the same transformations.

\section{Full-frame \corrupt{} examples}
\label{app:corruptionfigs}

\begin{figure}[p]
    \centering
    \includegraphics[width=0.98\linewidth]{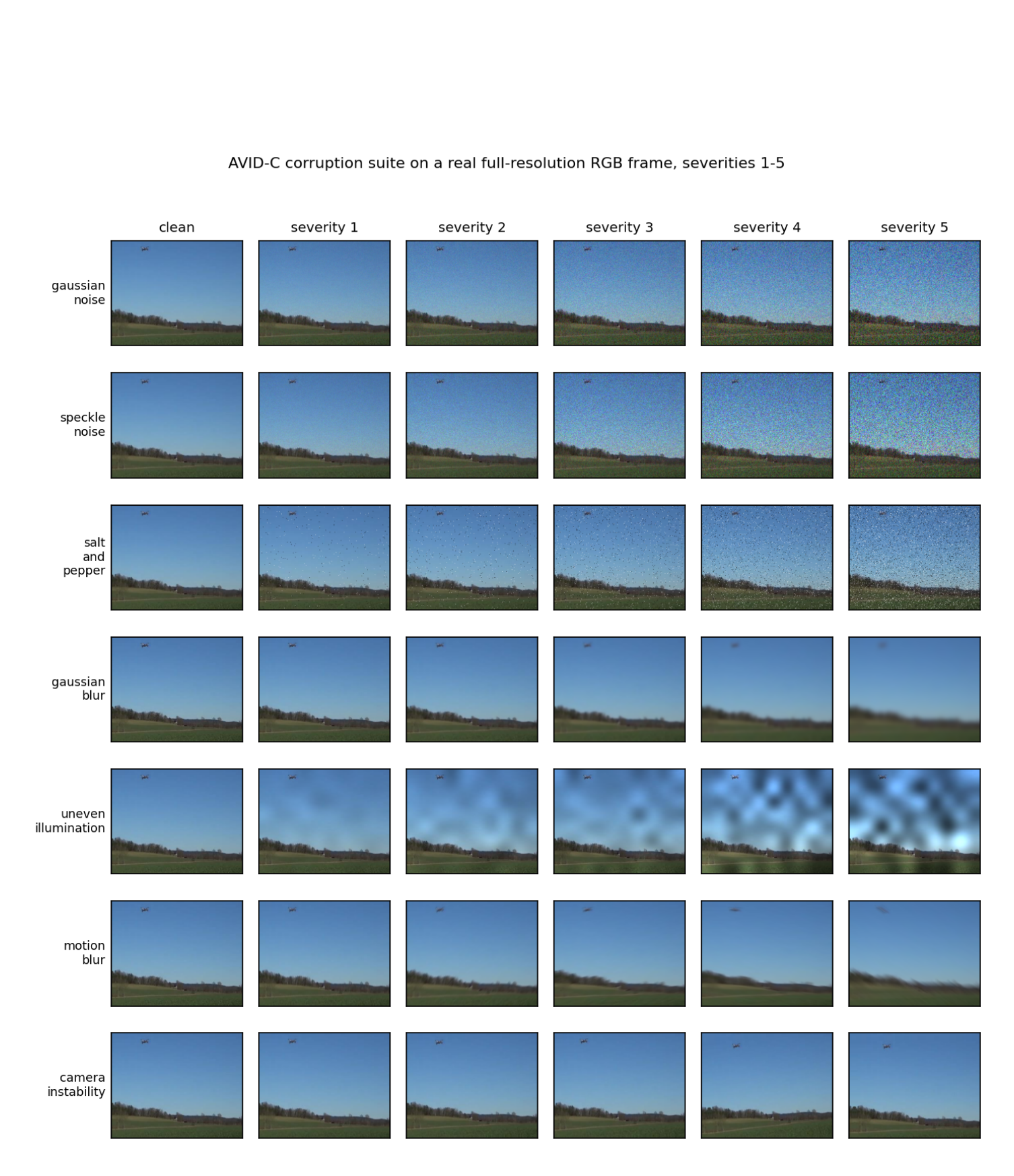}
    \caption{\textbf{Examples of simulated image degradation in RGB frames.}
    Each row shows one of the seven perturbation families in \corrupt{}, with severity increasing from left to right. The corresponding parameter values are listed in Table~\ref{tab:corruptions}.}
    \label{fig:rgbcorruptgrid}
\end{figure}

\begin{figure}[p]
    \centering
    \includegraphics[width=0.98\linewidth]{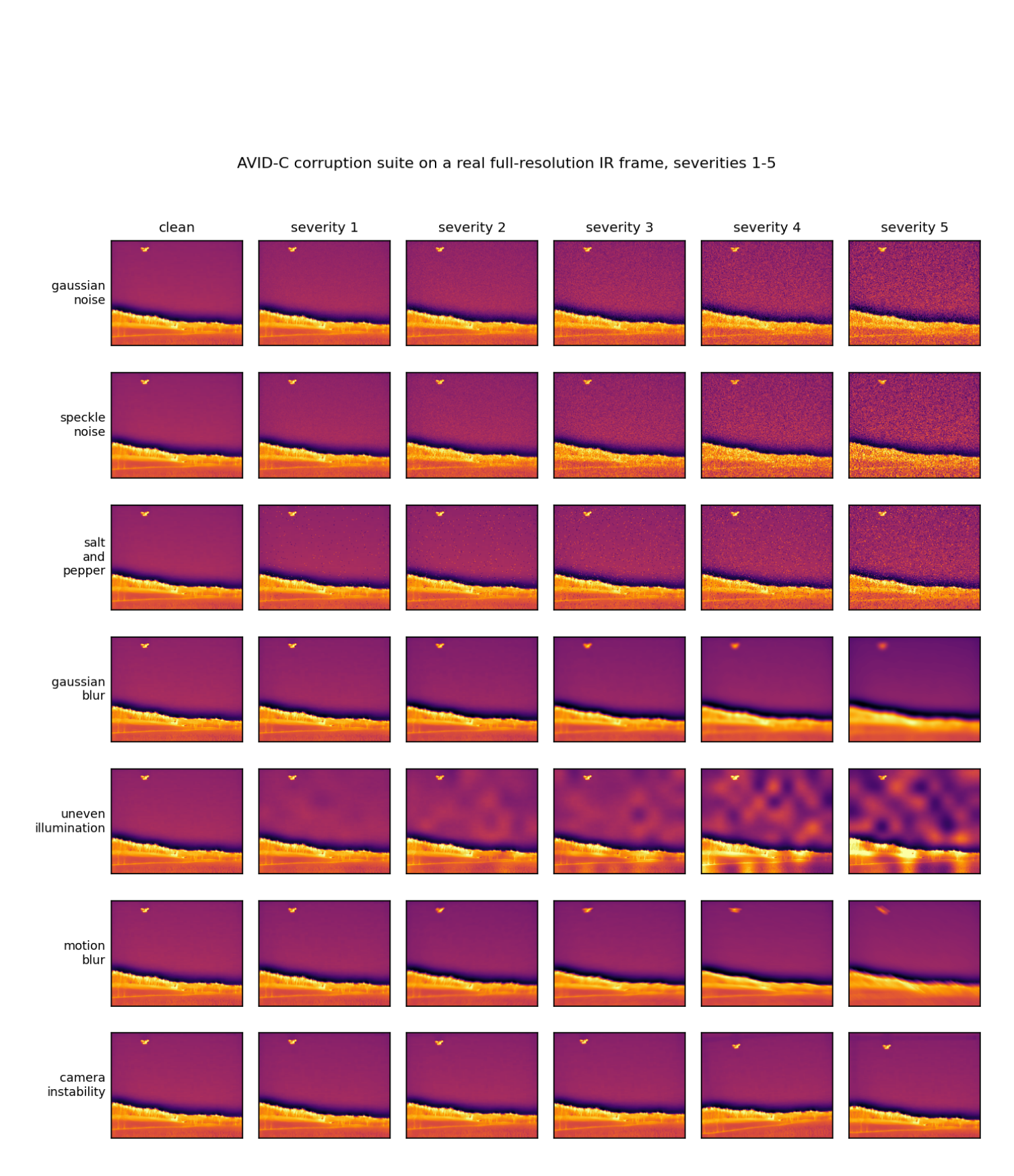}
    \caption{\textbf{Examples of simulated image degradation in infrared frames.}
    Each row shows one of the seven perturbation families across increasing severity levels, applied to the infrared imagery. The perturbation settings correspond to those used for the RGB examples.}
    \label{fig:ircorruptgrid}
\end{figure}

Figures~\ref{fig:rgbcorruptgrid} and \ref{fig:ircorruptgrid} show the seven \corrupt{} families at five severity levels in representative full-frame RGB and infrared observations.

The native-resolution examples make the transformations visible in the context of the small target and its background. Their parameter values are given in Table~\ref{tab:corruptions}.

RGB and infrared examples also illustrate that the same simulated transformation can alter different image structures in the two modalities.

\section{Additional dataset examples}
\label{app:diversity}

\begin{figure}[h]
    \centering
    \includegraphics[width=0.90\linewidth]{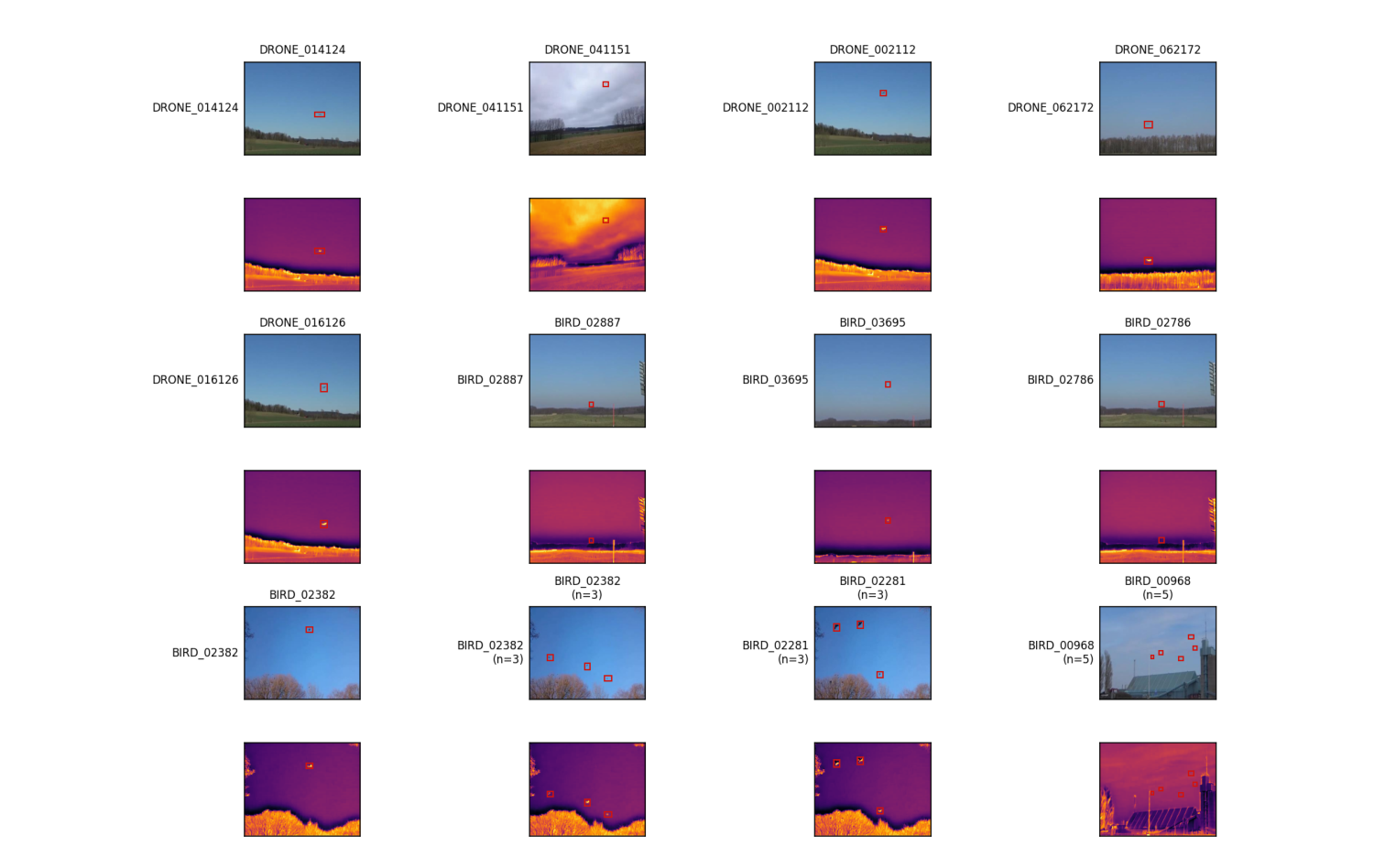}
    \caption{\textbf{Additional RGB-infrared examples from \dataset{}.}
    The montage includes paired frames from different source videos, showing variation in background, target appearance, target count, and thermal contrast. Red bounding boxes indicate the annotated targets.}
    \label{fig:diversity}
\end{figure}

Figure~\ref{fig:diversity} samples corresponding RGB-infrared observations from distinct source videos, illustrating variation in target appearance, background, target count, and thermal contrast.

This variation contextualizes the geometric and class-overlap statistics, while remaining limited to the surveillance conditions represented by the 115 source sequences.

\end{document}